\documentclass{article}
\PassOptionsToPackage{numbers, compress}{natbib}
\usepackage[preprint]{neurips_2025}
\usepackage{fix-cm}

\usepackage{xcolor}         
\usepackage{graphicx}
\usepackage{xspace}
\definecolor{linkColor}{rgb}{0.2,0.4,0.6}
\usepackage[utf8]{inputenc} 
\usepackage[T1]{fontenc}    
\usepackage[colorlinks=true,linkcolor=linkColor,citecolor=linkColor,filecolor=linkColor,urlcolor=linkColor]{hyperref}       
\usepackage{url}            
\usepackage{booktabs}       
\usepackage{amsfonts}       
\usepackage{stmaryrd}       
\usepackage{nicefrac}       
\usepackage{microtype}      

\usepackage{graphicx}
\usepackage{arydshln}
\usepackage{booktabs}
\usepackage{multirow}
\usepackage{caption}
\usepackage{subcaption}
\usepackage{makecell}
\usepackage{epigraph}
\usepackage{booktabs} 

\usepackage{CJKutf8}
\usepackage[utf8]{inputenc} 
\usepackage[T1]{fontenc}    
\usepackage{hyperref}       
\usepackage{url}            
\usepackage{booktabs}       
\usepackage{amsfonts}       
\usepackage{nicefrac}       
\usepackage{microtype}      
\usepackage{xcolor}         
\usepackage{markdown}

\usepackage{CJKutf8}
\usepackage{graphicx}
\usepackage{booktabs} 
\usepackage{amsmath,amsfonts}
\usepackage{algorithmic}
\usepackage{algorithm}
\usepackage{array}
\usepackage{float}
\usepackage{xcolor}
\usepackage{multirow}

\usepackage{tabularx}   
\usepackage{booktabs}   
\usepackage{url}        
\usepackage{ragged2e}   
\usepackage{geometry}

\usepackage{xcolor}
\usepackage{tcolorbox}

\usepackage{graphicx}

\usepackage{xcolor}
\usepackage{graphicx}
\newtcolorbox{markdownBox}{
    colback=gray!20,
    colframe=black,
    breakable,
    arc=3mm,
    title={},
    toptitle=0mm,
    bottomtitle=0mm,
    colbacktitle=gray!20,
    coltitle=black,
    fonttitle=\bfseries,
}

\definecolor{abstractbg}{gray}{0.95}
\makeatletter
\renewenvironment{abstract}{%
  \vskip 0.1in
  \begin{tcolorbox}[
    colback=abstractbg, colframe=abstractbg,
    arc=3mm, boxrule=0pt,
    left=6mm, right=6mm, top=4mm, bottom=4mm
  ]
}{%
  \end{tcolorbox}
}
\makeatother

\usepackage{csquotes}

\RequirePackage{algorithm}
\RequirePackage{algorithmic}

\usepackage{amsmath,amsfonts,bm}

\def\eqref#1{equation~\ref{#1}}

\def\1{\bm{1}}

\def\vh{{\bm{h}}}

\def\vs{{\bm{s}}}

\def\vx{{\bm{x}}}

\def\vz{{\bm{z}}}

\DeclareMathAlphabet{\mathsfit}{\encodingdefault}{\sfdefault}{m}{sl}
\SetMathAlphabet{\mathsfit}{bold}{\encodingdefault}{\sfdefault}{bx}{n}

\newcommand\our{\textsc{VibeVoice}}

\newcommand\sigvae{$\sigma$-VAE}

\newcommand{\huggingface}{\raisebox{-1.5pt}{\includegraphics[height=1.05em]{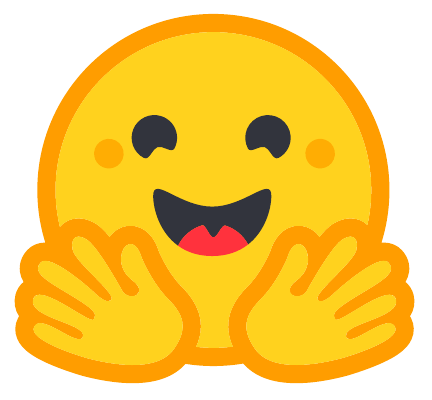}}\xspace}
\newcommand{\github}{\raisebox{-1.5pt}{\includegraphics[height=1.05em]{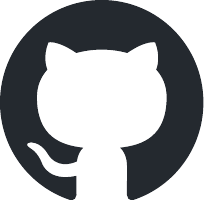}}\xspace}
\newcommand{\microphone}{{\raisebox{-1.5pt}{\includegraphics[height=1.05em]{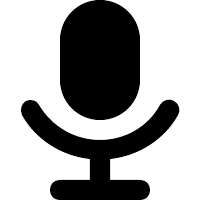}}}\xspace}
\newcommand{\house}{{\raisebox{-1.5pt}{\includegraphics[height=1.05em]{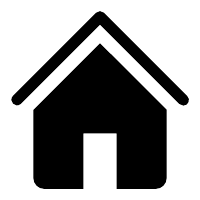}}}\xspace}

\title{\our{} Technical Report
}

\author{
Zhiliang Peng\thanks{~Core contributors. $\diamond$ Contact person: \href{mailto:fuwei@microsoft.com}{fuwei@microsoft.com}.},~~Jianwei Yu\footnotemark[1],~~Wenhui Wang\footnotemark[1],~~Yaoyao Chang\footnotemark[1],~~Yutao Sun\footnotemark[1],~~Li Dong\footnotemark[1] \\  
~\bf Yi Zhu, Weijiang Xu, Hangbo Bao, Zehua Wang, Shaohan Huang, Yan Xia, Furu Wei$^{\diamond}$ \\
Microsoft Research \\
~{\href{https://aka.ms/GeneralAI}{https://aka.ms/GeneralAI}}
}

\begin{document}

\maketitle



\begin{abstract}
This report presents \our{}, a novel model designed to synthesize long-form speech with multiple speakers by employing next-token diffusion~\cite{latentlm}, which is a unified method for modeling continuous data by autoregressively generating latent vectors via diffusion. To enable this, we introduce a novel continuous speech tokenizer that, when compared to the popular Encodec model, improves data compression by 80 times while maintaining comparable performance. The tokenizer effectively preserves audio fidelity while significantly boosting computational efficiency for processing long sequences. Thus, \our{} can synthesize long-form speech for up to 90 minutes (in a 64K context window length) with a maximum of 4 speakers, capturing the authentic conversational ``vibe'' and surpassing open-source and proprietary dialogue models.




\end{abstract}

\begin{table}[H]
\centering
\begin{tabular}{@{}r@{\hspace{2pt}}l@{}} 
\house & \textbf{Project Page}: \href{https://microsoft.github.io/VibeVoice}{\texttt{aka.ms/VibeVoice}}\\
\github & \textbf{Code}: \href{https://github.com/microsoft/VibeVoice}{\texttt{github.com/microsoft/VibeVoice}} \\
\huggingface & \textbf{Hugging Face}: 
\href{https://huggingface.co/collections/microsoft/vibevoice-68a2ef24a875c44be47b034f}{\texttt{microsoft/VibeVoice}} \\
\microphone & \textbf{Demo}: 
\href{https://aka.ms/VibeVoice-Demo}{\texttt{aka.ms/VibeVoice-Demo}}\\
\end{tabular}
\end{table}

\begin{figure}[!h]
\centering
\includegraphics[width=\linewidth]{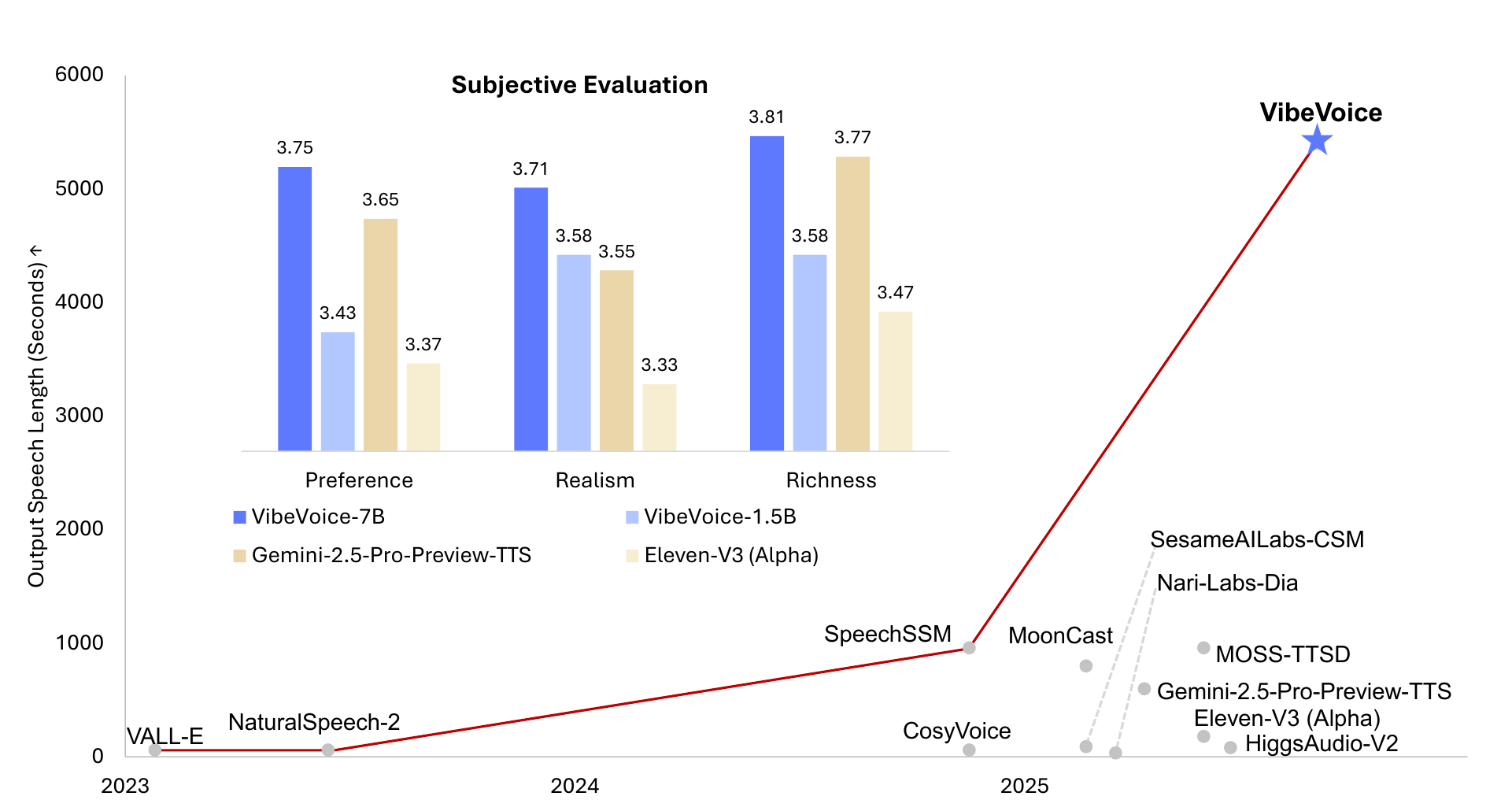}

  
\caption{
\our{} is capable of synthesizing 5,000+ seconds of audio while consistently outperforming strong open/closed-source systems in subjective evaluations of preference, realism, and richness.
}
\label{fig:result}
\end{figure}

\section{Introduction}

While recent advancements in Text-to-Speech (TTS) synthesis have achieved remarkable success in generating high-fidelity, natural-sounding speech for single speakers in short utterances~\cite{valle, shen2023naturalspeech, anastassiou2024seedtts,voicebox,chen2024f5,du2024cosyvoice2,jia2025ditar,ye2025llasa}, a significant frontier remains in the scalable synthesis of long-form, multi-speaker conversational audio, such as podcasts and multi-participant audiobooks. Although traditional systems can technically produce such audio by concatenating individually synthesized utterances, achieving natural turn-taking and content-aware generation are major challenges. Recently, research on multi-speaker long conversational speech generation has begun to emerge~\cite{NotebookLM, park2024speechssm, narilabs-dia, Moss-TTSD, SesameAILabs-csm, liu2024autoregressive}. However, most of these works are either not open-sourced~\cite{NotebookLM, park2024speechssm} or still face challenges in terms of generation length and stability~\cite{zhang2025covomix2, SesameAILabs-csm, mooncast, Moss-TTSD}.  

\begin{figure}[t]
\centering
\includegraphics[width=1\linewidth]{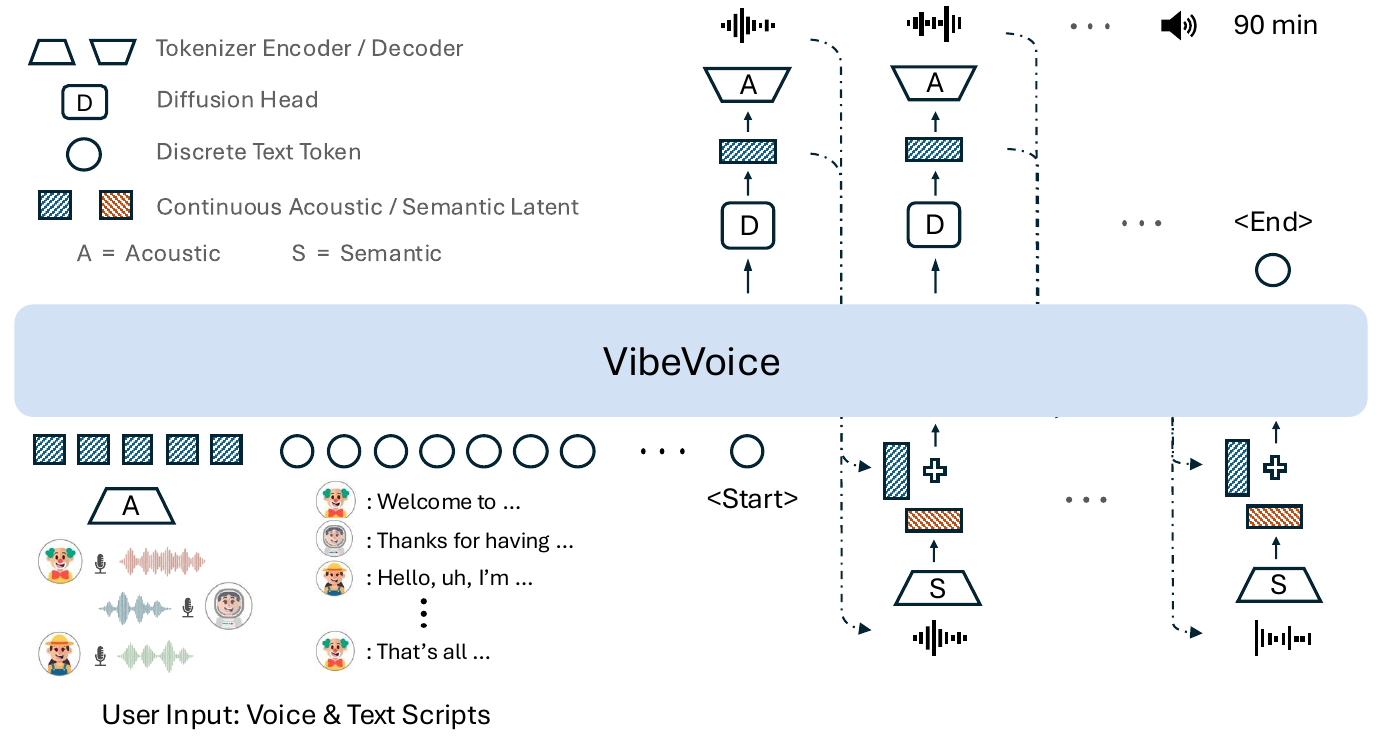}
\caption{
\our{} employs next token diffusion framework as in LatentLM~\cite{latentlm} to synthesize long-form and multi-speaker audios.
Voice prompts and text scripts provide initial input. \our{} processes hybrid context features, and its hidden states condition a token level Diffusion Head (D), which predicts acoustic VAE for speech segments, subsequently recovered by acoustic decoder (A).
}
\label{fig:vibepod_inf_pip}
\end{figure}

In this work, we introduce \our{}, as illustrated in Figure~\ref{fig:vibepod_inf_pip}, a novel framework developed for the scalable synthesis of long-form and multi-speaker speech. 
To support long audio generation, we have pioneered the development of a causal speech tokenizer that achieves a 3200$\times$ compression rate (i.e., 7.5~Hz frame rate). 
In our experiments, this highly efficient tokenizer maintains a speech-to-text token ratio of approximately 2:1, meaning two speech tokens are roughly equivalent to one BPE~\cite{bpe} text token. 

We utilize a pre-trained Large Language Model (LLM, e.g., Qwen2.5~\cite{qwen2_5}) to interpret complex user inputs, including detailed text sentences and role assignments. 
We have streamlined the architecture by removing unnecessary prior designs: voice latent features and text scripts are concatenated into a single sequence and fed directly into the LLM. 
The LLM then processes this context to predict a hidden state, which in turn conditions a lightweight, token-level Diffusion Head~\cite{mar}. 
This diffusion head is responsible for predicting the continuous Variational Autoencoder (VAE) features, which are subsequently recovered into the final audio output by speech tokenizer decoder.

Despite its architectural simplicity, \our{} yields an exceptionally powerful TTS model.
It demonstrates remarkable flexibility in handling multiple speakers and achieves a synthesis length of up to 90 minutes. 
Scaling the LLM from 1.5B to 7B, the larger model exhibits significant gains in perceptual quality, delivering richer timbre, more natural intonation, and enhanced transfer capabilities, such as in cross-lingual applications.

\section{Method}


\subsection{Speech Tokenizers}\label{sec:tokenizer}

We employ two separate tokenizers as input to learn both acoustic and semantic features. In our experiments, generating long-form speech benefits from this separate design.

\textbf{Acoustic Tokenizer} adopts the principles of a Variational Autoencoder (VAE)~\cite{vae}, specifically drawing inspiration from the \sigvae{} variant proposed in LatentLM~\cite{latentlm} to mitigate potential variance collapse issues of VAEs when used in autoregressive modeling settings. The process involves an encoder network, parameterized by $\phi$, which maps the input audio $\vx$ to the parameters of a latent distribution, primarily the mean $\mu$. 
Notably, variance $\sigma$ is a pre-defined distribution ($\mathcal{N}(0, C_{\sigma}$)) in \sigvae{}, rather than a learnable distribution in VAE~\cite{vae}.
A latent vector $\vz$ is then sampled using the reparameterization trick. Following the \sigvae{} approach to ensure robust variance for autoregressive modeling, we can formulate this as: $\vz = \mu + \sigma \odot \boldsymbol{\epsilon}, \text{where} ~ \boldsymbol{\epsilon} \sim \mathcal{N}(0, 1), ~ \sigma \sim \mathcal{N}(0, C_{\sigma})$.

The architecture is a mirror-symmetric encoder-decoder structure. The encoder employs a hierarchical design with 7 stages of modified Transformer blocks \cite{transformer} (using 1D depth-wise causal convolutions instead of self-attention module) for efficient streaming processing. Six downsampling layers achieve a cumulative 3200X downsampling rate from a 24kHz input, yielding 7.5 tokens/frames per second. Each encoder/decoder component has approximately 340M parameters. The training objective follows the DAC~\cite{dac}, including its discriminator and loss designs.
 
\textbf{Semantic Tokenizer} mirrors the hierarchical architecture of the Acoustic Tokenizer's encoder, but without VAE components, as its objective is deterministic content-centric feature extraction. The main difference is the training objective, which uses Automatic Speech Recognition (ASR) as the proxy task. 
During training, its output is decoded by several Transformer decoder layers to predict text transcripts, aligning the semantic encoder's representations with textual semantics. This decoder is discarded after pre-training.

\subsection{\our{}}

\our{} employs a Large Language Model (LLM) as its core sequence model, integrated with specialized audio encoding and diffusion-based decoding modules to achieve scalable, high-fidelity multi-speaker speech synthesis. 
The overall inference architecture is depicted in Figure~\ref{fig:vibepod_inf_pip}. 

\textbf{Input Representation}: The model input $X$ is formed by concatenating the voice font features and the text script embeddings, specified by users, interleaved with role identifiers ($\texttt{Speaker}_k$): $X = [\texttt{Speaker}_1: \vz_{1}, \texttt{Speaker}_2: \vz_{2}, ..., \texttt{Speaker}_N: \vz_{N}] + [\texttt{Speaker}_1: T_1, \texttt{Speaker}_2: T_2, ..., \texttt{Speaker}_N: T_N]$, where $\vz_{N}$ is acoustic latent representations and $T_N$ is each role's text scripts. 
For the generated speech segment $\vs$, it will be encoded by acoustic tokenizer and semantic tokenizer to form the hybrid speech representation for the auto-regressive modeling.

\textbf{Token-Level Diffusion}:
To synthesize speech in a streaming way, \our{} employs a lightweight diffusion head~\cite{mar} conditioned on the LLM's hidden state of each token, $\vh_i$.
During training, this diffusion head is optimized to reverse a forward noising process by predicting the noise~\cite{ddpm} added to the clean acoustic VAE features $z_{a,i}$.
During inference, this diffusion head iteratively refines a randomly sampled Gaussian noise vector to predict the target acoustic VAE feature, $\vz_{a,i}$.
This denoising process is enhanced using Classifier-Free Guidance (CFG), which interpolates between a conditional prediction (guided by $\vh_i$) and an unconditional prediction. 
An efficient sampler, such as DPM-Solver++~\cite{dpm-solver,dpm-solver++}, is utilized to accelerate this iterative process, ultimately yielding a clean acoustic feature estimate.

We instantiated \our{}'s core LLM using the 1.5B and 7B parameter versions of Qwen2.5 \cite{qwen2_5}. The diffusion head \cite{mar} comprises 4 layers. 
During \our{} training, the pre-trained acoustic and semantic tokenizers remained frozen, with only the LLM and diffusion head parameters being learnable. We employed a curriculum learning strategy for the LLM input sequence length, progressively increasing from 4,096 to 65,536 tokens. 
The guidance scale is 1.3 and the iterative denoising step is 10 for \our{}.

\section{Results}

\subsection{\our{} Podcast}

\begin{table*}[t!]
\centering
\newcommand{\err}[1]{\tiny$\pm$#1}
\setlength{\tabcolsep}{6pt} 
\resizebox{\linewidth}{!}{%
\begin{tabular}{@{}lcccccccc@{}}
\toprule
\multirow{2}{*}{\textbf{Model}}& \multicolumn{4}{c}{\textbf{Subjective}} & \multicolumn{3}{c}{\textbf{Objective}} \\
\cmidrule(lr){2-5} \cmidrule(lr){6-8}
& \textbf{Realism} & \textbf{Richness} & \textbf{Preference} & \textbf{Average} & \textbf{WER (Whisper)} & \textbf{WER (Nemo)} & \textbf{SIM} \\
\midrule
Nari Labs Dia~\cite{narilabs-dia} & - & - & - & - & 11.96 & 10.79 & 0.541 \\
Mooncast~\cite{mooncast} & - & - & - & - & 2.81 & 3.29 & 0.562 \\
SesameAILabs-CSM~\cite{SesameAILabs-csm} & 2.89 \err{1.15} & 3.03 \err{1.11} & 2.75 \err{1.08} & 2.89 \err{1.12} & 2.66 & 3.05 & 0.685 \\
Higgs Audio V2~\cite{higgsaudio2025} & 2.95 \err{1.13} & 3.19 \err{1.06} & 2.83 \err{1.16} & 2.99 \err{1.13} & 5.94 & 5.97 & 0.543 \\
Elevenlabs v3 alpha~\cite{elevenlabs-v3} & 3.34 \err{1.11} & 3.48 \err{1.05} & 3.38 \err{1.12} & 3.40 \err{1.09} & 2.39 & 2.47 & 0.623 \\
Gemini 2.5 pro preview tts~\cite{gemini-tts} & 3.55 \err{1.20} & 3.78 \err{1.11} & 3.65 \err{1.15} & 3.66 \err{1.16} & 1.73 & 2.43 & - \\
\midrule
\our{}-1.5B & 3.59 \err{0.95} & 3.59 \err{1.01} & 3.44 \err{0.92} & 3.54 \err{0.96} & \textbf{1.11} & \textbf{1.82} & 0.548 \\
\our{}-7B & \textbf{3.71} \err{0.98} & \textbf{3.81} \err{0.87} & \textbf{3.75} \err{0.94} & \textbf{3.76} \err{0.93} & 1.29 & 1.95 & \textbf{0.692} \\
\bottomrule
\end{tabular}%
}
\caption{Human subjective and objective evaluation results. For all subjective metrics and SIM-O, higher scores are better. For WER, lower scores are better. Best results are in \textbf{bold}.}
\label{tab:combined_evaluation}
\end{table*}

We conducted both objective and subjective evaluations to benchmark the performance of the proposed \our{} against recent state-of-the-art conversational speech generation systems~\cite{narilabs-dia, mooncast, SesameAILabs-csm, higgsaudio2025, elevenlabs-v3, gemini-tts}.

To manage the labor-intensive and time-consuming nature of subjective evaluation, we designed a compact test set. This set consists of 8 long conversational transcripts with a total duration of about 1 hour. We used speech prompts to ensure consistent timbre across the different models. Since Gemini 2.5 Pro preview TTS does not support speech-prompt control, we used its default male and female voices for comparison instead.

For our objective evaluation, we measure Word Error Rate (WER) and speaker similarity. WER is obtained by transcribing the generated speech using Whisper-large-v3~\cite{whisper} and Nemo ASR~\cite{xu2023nemo}. Speaker similarity (SIM) is computed by extracting speaker embeddings with WavLM-large~\cite{chen2022wavlm}.  

For subjective evaluation, we recruited 24 human annotators to provide Mean Opinion Scores (MOS) across three dimensions: \textbf{Realism} (how natural and human-like the speech sounds, including prosody, emotion, and the smoothness of speaker turns), \textbf{Richness} (the expressiveness of the speech in terms of tone and emotion, including variation and adaptation to context), and \textbf{Preference} (overall listener enjoyment and subjective preference, reflecting naturalness, pleasantness, and engagement). The evaluation covered six models with all eight test samples, meaning that each annotator listened to approximately six hours of audio in total.

We can observe that: \textbf{The proposed \our{} models outperform all other top-tier models on long conversational speech generation across both objective and subjective metrics.}  Compared with the \our{}-1.5B model, the \our{}-7B model achieves significantly better performance on all objective metrics and SIM, while maintaining a comparable WER.

\subsection{\our{} Short Utterance}

We evaluate \our{} on the SEED test sets~\cite{seedtts}, a widely used benchmark composed of short utterances.
For evaluation, approximately 1,000 English samples and 2,000 Chinese samples are drawn from the CommonVoice dataset, denoted as \emph{test-en} and \emph{test-zh}, respectively.
We compute word error rate (WER) using Whisper-large-v3 for \emph{test-en} and Paraformer~\cite{paraformer} for \emph{test-zh}. 
For speaker similarity (SIM), we adopt a WavLM-large~\cite{chen2022wavlm} model.

Table~\ref{tbl:eval_seedtts} presents the results on the SEED test sets.
Although our model is primarily trained on long-form speech, it demonstrates strong generalization on short-utterance benchmarks.
In addition, by employing a lower frame rate, our model substantially reduces the number of decoding steps required to synthesize one second of speech.

\begin{table*}[t]
\centering
\small
\begin{tabular}{lc|cc|cc}
\toprule
\multirow{2}{*}{\textbf{Model}} & \multirow{2}{*}{\textbf{Frame Rate}} & \multicolumn{2}{c|}{\textbf{test-zh}} & \multicolumn{2}{c}{\textbf{test-en}} \\
& & \textbf{CER(\%)~$\downarrow$} & \textbf{SIM~$\uparrow$} & \textbf{WER(\%)~$\downarrow$} & \textbf{SIM~$\uparrow$} \\
\midrule
MaskGCT~\cite{maskgct} & 50 & 2.27 & 0.774 & 2.62 & 0.714 \\
Seed-TTS~\cite{seedtts} & - & 1.12 & 0.796 & 2.25 & 0.762 \\
FireRedTTS~\cite{fireredtts} & 25 & 1.51 & 0.635 & 3.82 & 0.460 \\
CosyVoice 2~\cite{cosyvoice2} & 25 & 1.45 & 0.748 & 2.57 & 0.652 \\
Spark TTS~\cite{sparktts} & 50 & 1.20 & 0.672 & 1.98 & 0.584 \\
\midrule
\our{}-1.5B & 7.5 & 1.16 & 0.744 & 3.04 & 0.689 \\
\bottomrule
\end{tabular}
\caption{Results on the SEED test sets.}
\label{tbl:eval_seedtts}
\end{table*}

\begin{table*}[t]
\centering
\resizebox{.9\textwidth}{!}{
\begin{tabular}{lcc|ccc|ccc}
\toprule
\multirow{2}{*}{\textbf{Tokenizer}} & \multirow{2}{*}{\textbf{$N_q$}} & \bf Token & \multicolumn{3}{c|}{\textbf{test-clean}} & \multicolumn{3}{c}{\textbf{test-other}} \\
& & \bf Rate & \bf PESQ & \bf STOI & \bf UTMOS & \bf PESQ & \bf STOI & \bf UTMOS \\
\midrule
Ground-Truth & - & - & - & - & 4.056 & - & - & 3.483 \\
Encodec~\cite{encodec} & 8 & 600 & 2.72 & \bf 0.939 & 3.04 & 2.682 & \bf 0.924 & 2.657 \\
DAC~\cite{dac} & 4 & 400 & 2.738 & 0.928 & 3.433 & 2.595 & 0.908 & 2.945 \\
Encodec~\cite{encodec} & 4 & 300 & 2.052 & 0.901 & 2.307 & 2.052 & 0.884 & 2.088 \\
SpeechTokenizer~\cite{zhang2023speechtokenizer} & 4 & 300 & 1.931 & 0.878 & 3.563 & 1.737 & 0.837 & 3.018  \\
DAC~\cite{dac} & 1 & 100 & 1.246 & 0.771 & 1.494 & 1.245 & 0.751 & 1.499 \\
WavTokenizer~\cite{wavtokenizer} & 1 & 75 & 2.373 & 0.914 & 4.049 & 2.261 & 0.891 & 3.431 \\
WavTokenizer~\cite{wavtokenizer} & 1 & 40 & 1.703 & 0.862 & 3.602 & 1.662 & 0.834 & 3.055 \\ 
\midrule
Ours (Acoustic) & 1 & 7.5 & \bf 3.068 & 0.828 & \bf 4.181 & \bf 2.848 & 0.823 & \bf 3.724 \\
\bottomrule
\\
\end{tabular}
}
\vspace{-1em}
\caption{Objective evaluation of speech tokenizer’s reconstruction quality on the LibriTTS test-clean and test-other datasets. $N_q$ denotes the number of quantizers (VAE for us). Token Rate indicates the number of tokens/frames generated per second of audio. Higher PESQ, STOI, and UTMOS scores indicate better performance. Best results are in \textbf{bold}.}
\label{tbl:eval_tok}
\end{table*}

\subsection{Tokenizer Reconstruction}

The fidelity of audio reconstructed from acoustic tokens is a critical indicator of the tokenizer's efficacy in preserving essential acoustic information, particularly under high compression rates. To quantify this, we measured PESQ \cite{pesq}, STOI \cite{stoi} and UTMOS \cite{utmos} on both the LibriTTS test-clean and test-other datasets \cite{libritts}. 
Table~\ref{tbl:eval_tok} shows that our acoustic tokenizer, uniquely operating at an ultra-low 7.5~Hz, achieves leading PESQ and UTMOS scores on both test-clean (PESQ: 3.068, UTMOS: 4.181) and test-other (PESQ: 2.848, UTMOS: 3.724) subsets. This demonstrates its capacity for high-fidelity, perceptually excellent audio reconstruction despite aggressive compression, which is a key factor for \our{}'s scalability with long-form audio.

\section{Conclusion, Limitations, and Risks}

We introduced \our{}, a novel framework for long-form and multi-speaker speech generation. By integrating efficient hybrid speech representations from specialized ultra-low frame rate (7.5 Hz) acoustic and semantic tokenizers with an end-to-end LLM-based next-token diffusion framework, \our{} achieves state-of-the-art performance. It scalably synthesizes high-quality audio for up to 90 minutes with up to 4 speakers, demonstrably surpassing existing baselines in both subjective perceptual quality—including preference, realism, and richness—and objective metrics like WER, thereby significantly advancing the capabilities of conversational TTS.

English and Chinese only: Transcripts in languages other than English or Chinese may result in unexpected audio outputs. 

Non-Speech Audio: The model focuses solely on speech synthesis and does not handle background noise, music, or other sound effects. 

Overlapping Speech: The current model does not explicitly model or generate overlapping speech segments in conversations.

Potential for Deepfakes and Disinformation: High-quality synthetic speech can be misused to create convincing fake audio content for impersonation, fraud, or spreading disinformation. Users must ensure transcripts are reliable, check content accuracy, and avoid using generated content in misleading ways. 

We do not recommend using \our{} in commercial or real-world applications without further testing and development. This model is intended for research and development purposes only. Please use responsibly.

\bibliography{vibepod}
\bibliographystyle{alpha}

\end{document}